# Binary Space Partitioning as Intrinsic Reward

Wojciech Skaba

AGINAO, Trubadurow 11, 80205 Gdansk, Poland
`wojciech.skaba@aginao.com`

**Abstract.** An autonomous agent embodied in a humanoid robot, in order to learn from the overwhelming flow of raw and noisy sensory, has to effectively reduce the high spatial-temporal data dimensionality. In this paper we propose a novel method of unsupervised feature extraction and selection with binary space partitioning, followed by a computation of information gain that is interpreted as intrinsic reward, then applied as immediate-reward signal for the reinforcement-learning. The space partitioning is executed by tiny codelets running on a simulated Turing Machine. The features are represented by concept nodes arranged in a hierarchy, in which those of a lower level become the input vectors of a higher level.

## 1 Introduction

For an autonomous humanoid robot, learning a cognitive model from the natural environment, there seems to be no direct correspondence between low level sensory and high level external motivation. Furthermore, the reinforcement-learning becomes ineffective if the extrinsic reward signal propagates through too many states. The learning could be improved, however, should a good candidate for the immediate reward be found.

The term *intrinsic motivation* was borrowed by cognitive scientists from the psychology, to mean that an agent is engaged in some activity for its own sake, possibly activity in taking pleasure, rather than working to fulfil some external drives. This motivational force is referred to as independent ego-energy, based in organism needs to be competent and self-determining [1]. Closely related is the term *intrinsic reward*, to mean a reinforcement stimulus of the intrinsic motivation.

As for studies on epigenetic robotics and autonomous agents, however, the term has been conceptualized differently and in many distinct ways, while a unified definition seems not to exist yet [2]. Miscellaneous measures of intrinsic motivation have been proposed, including: information gain, curiosity, novelty, prediction error, competence progress, relative entropy, compression progress, etc. [2], [3]. We will focus here on a purely information-theoretic based approach, where intrinsic reward is defined as averaged information gain assigned to an action taken to evaluate a candidate feature. The action to be taken is represented by a directed edge connecting two con-

cept nodes, the head to stand for the feature under consideration, the tail to stand for
the input vector, possibly a lower level feature selected earlier, or raw sensory data
(also represented by a feature vector). The intrinsic reward is then applied as immediate-reward signal for the reinforcement-learning algorithm, becoming the only source
of reward. The latter to mean that no external motivation driven reward is employed.

## 2      AGINAO Cognitive Architecture

AGINAO is a project to build a human-level artificial general intelligence (AGI) system by embodiment of the cognitive engine in the NAO humanoid robot. It was first
introduced in [4]. A more detail presentation of the self-programming engine is given
in [5]. The open-ended learning, executed by the cognitive engine, is a result of a
fully automatic self-programming development of a hierarchy of interconnected *concepts*, as shown in Fig. 1.

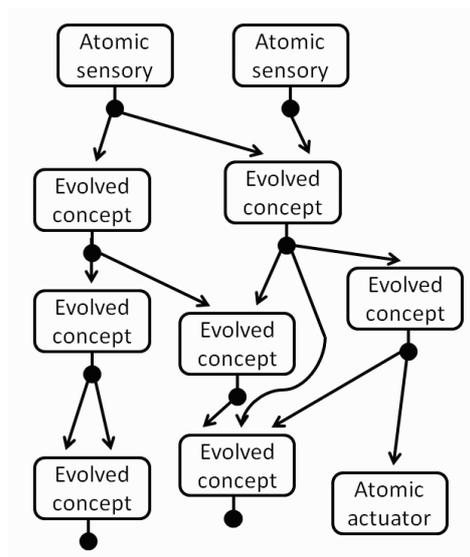

**Fig. 1.** Sample concept network.

The evolved concept nodes represent the features of the spatial-temporal sensory patterns of the natural world and are evaluated concurrently at all levels of the hierarchy.
The predefined atomic sensory and actuator concepts stand for the root and terminal
nodes, respectively. An output of a concept becomes an input of a higher level concept, so the distinction between features and a patterns disappears. There are also
concepts that act as the procedural memory, and concepts that behave like functions.
For that reason, we prefer the word *pattern* to name the entities being discovered and
managed by the concepts, while the word *feature* may be used for better communication, where applicable. A patternist philosophy of mind is thoroughly discussed in [6].

The idea that a single algorithm may be used to process both the spatial and temporal aspects of a pattern, and that pattern processing should be conducted simultaneously at all levels of the hierarchy, is presented in [7].

The feature detection is performed by a tiny piece of machine code, a *codelet*, embedded in a concept node. Following the execution, the action values and other parameters that govern the dynamic structure of the network are updated, effectively to mean feature extraction and selection. The codelets consist of instructions of a custom-designed virtual machine, a simulated Turning Machine. A typical setting of a 2-input concept is shown in Fig. 2. and a sample codelet in Fig. 3.

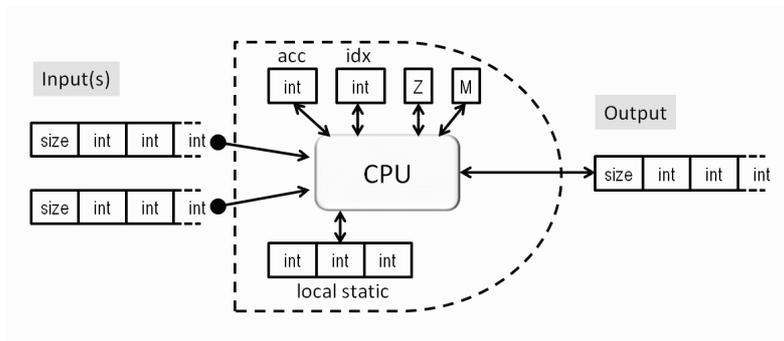

**Fig. 2.** A sample 2-input concept node and virtual processor.

```
0000 MOV A, var1[00]
0005 ADD A, var2[00]
0010 APPEND, A
0011 JZ 0005
0014 RET
```

**Fig. 3.** Sample program in machine code.

The basic internal type is a 16-bit integer (`int`). The information is exchanged between the concepts using a uniform data format, a vector of integers of know size. Fig. 4. depicts a sample feature vector of a visual pixel of the YUV color space.

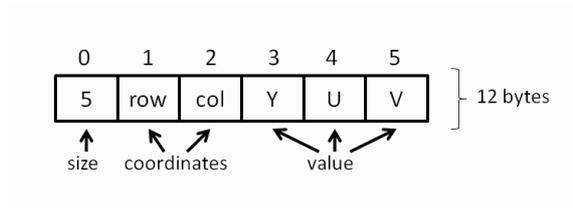

**Fig. 4.** Internal data format.

The input feature vector, as depicted in Fig. 2., consists of two features, the values being vectors too, possibly of other feature vectors. Should a feature be detected by the execution of the concept's codelet, the output would be a feature vector, too.

It must be highlighted, however, that concept network is not a neural network. The concepts and the links (edges) are stored in a depository and launched as individual threads. Multiple concurrent runtime-threads may coexist and be used to evaluate a single concept. The codelet code as such is generated in a process called *heuristic search in program space*, basically random, creating programs of highly non-linear behavior and of flexibility theoretically equivalent to a Universal Turing Machine.

Last but not the least, if two concepts are connected by an edge to mean an action, it is the head concept's codelet that is executed as an action, while the action values are stored in the tail concept(s), individually for each edge outgoing from the tail.

## 3  Intrinsic Motivation

Whether external drives are primary or secondary to a more basic motivational mechanism, it may be questioned [8]. For example, the hunger drive seems to be a very basic one for humans. There are historical examples, however, of people who have committed a sort of starvation suicide for some ideas. What follows, a stronger intrinsic motivational reward must exist, one that seems to be of *information seeking* type. The reward must originate from the internal model of the world, rather than from the external drives. It is conjectured here that, since virtually any drives may be overpowered by the intrinsic motivation, a cognitive model may be built with extrinsic rewards being only secondary to the intrinsic reward mechanism, and merely reflected there.

The proposed measure of intrinsic motivation is based on the notion of *self-information* [9] or *information surprisal* [10] associated with the execution of a concept's codelet, i.e. the amount of information provided by an event being a successful (pattern matching) execution of the codelet. The resulting intrinsic reward is the averaged self-information gain.

Let $\omega$ be the outcome of a random variable with probability $P(\omega)$. Then the self-information may be computed as

$$I(\omega) = -\log_2(P(\omega)) \qquad (1)$$

From now on, let us assume that each concept has only two outcomes, $\omega_{pos}$ for pattern matching and $\omega_{neg}$ for pattern not matching, and we appreciate the positive matches only. The more unique a pattern, the more information it entails, which reflects our intuition. We want our intrinsic reward mechanism to maximize the information gain in time, the time unit to be understood as a step of a (non-Markovian) decision process. Unfortunately, if the probability of an outcome decreases, we have to execute our concept codelet multiple times to get a match. Consequently, we get the following definition of intrinsic reward:

$$r(\omega_{pos}) = -P(\omega_{pos})\log_2(P(\omega_{pos})) \qquad (2)$$

Self-information is a special case of Kullback–Leibler distance from a Kronecker delta representing the matching pattern to the probability distribution:

$$I(\omega_{pos}) = \sum_{i=pos,neg} \delta_{pos,i} \log_2 \frac{\delta_{pos,i}}{P(\omega_i)} = 1 * \log_2 \frac{1}{P(\omega_{pos})} = -\log_2(P(\omega_{pos})) \quad (3)$$

An independent approach based on a measure of information gain calculated from the Kullback-Leibler distance was presented in [11].

The measure of self-information has an interesting additive property. If two independent events $A$ and $B$ with outcomes $\omega_A$ and $\omega_B$ have the probabilities $P(\omega_A)$ and $P(\omega_B)$, then the resulting information gain is

$$I(\omega_{A \cap B}) = I(\omega_A) + I(\omega_B) = -\log_2(P(\omega_A)) - \log_2(P(\omega_B)) \quad (4)$$

This may be a result of executing two concept codelets in sequence. It is quite likely, however, that a single concept codelet is performing exactly the same function as two concept in sequence, possibly created by a concatenation of the codelets. The resulting information gain would be

$$I(\omega_{A \cap B}) = -\log_2(P(\omega_A)P(\omega_B)) = -\log_2(P(\omega_A)) - \log_2(P(\omega_B)) \quad (5)$$

i.e. exactly what would be expected.

On the other hand, however, if the two concepts are executed in sequence[1], the resulting reward would be

$$r(\omega_{A \cap B}) = r(\omega_A) + r(\omega_B) = -P(\omega_A)\log_2(P(\omega_A)) - P(\omega_B)\log_2(P(\omega_B)) \quad (6)$$

while, the same function executed as a single concept would result in

$$r(\omega_{A \cap B}) = -P(\omega_A)P(\omega_B)\log_2(P(\omega_A)P(\omega_B)) \quad (7)$$

which is less than (6). This reflects the idea that getting the reward in separate steps is potentially more informative than doing everything in one step, especially in case when already

$$-P(\omega_A)\log_2(P(\omega_A)) > -P(\omega_A)P(\omega_B)\log_2(P(\omega_A)P(\omega_B)) \quad (8)$$

that hold if

$$P(\omega_A) < \frac{1}{e} \quad (9)$$

provided $A$ is the first step (see Fig.5.). Separate steps may be preferred because—after the first step was executed—we get more opportunities, while assuming a priori that a unique pattern will be encountered is always risky. One might even erroneously conclude that splitting concepts into separate steps is always beneficial when (9) doesn't hold, at least as long as (roughly):

$$P(\omega_B) < \gamma \qquad (10)$$

where $\gamma$ in the discount-factor of TD-learning. That's not observed in practical implementations, however. The computational overhead of executing two separate concepts instead of one must also be taken into account.

---

[1] A TD-learning discount factor $\gamma$, that would normally be included, is temporarily omitted.

The reward function is depicted on diagram in Fig. 5. and reaches the maximum at:

$$(-p \log_2 p)' = -\frac{\ln p}{\ln 2} - \frac{1}{\ln 2} = 0 \Rightarrow p = \frac{1}{e} \qquad (11)$$

What follows is that preferred would be concepts with the probability of a positive match around the reciprocal of $e$. This must not be confused with the mentioned above question of splitting the concepts, for in the former case the probability is given, while in the latter case we search the space of concepts with unknown probabilities.

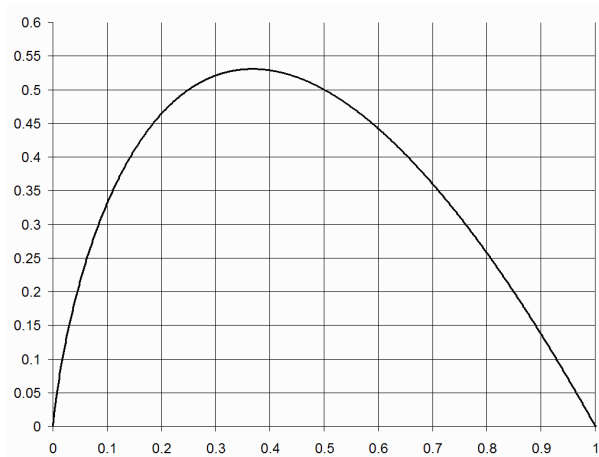

**Fig. 5.** $-p \log_2 p$

The properties of the reward function may be illustrated with the following example. Imagine, we want our cognitive engine to learn the most effective method for the graphical recognition of the 25 letters of the Latin alphabet. The simplest method would be to design an individual concept for each letter. Since, however, the individual probabilities of occurrence of each letter are rather low, so is the reward (left end of the diagram). We could increase the expected probability by designing an algorithm to exclude a letter rather than finding a match. Unfortunately, the resulting self-information gain would be rather low (right end).

An alternative approach could be based on detecting some feature first, like finding whether the observed letter contains a vertical bar. Some 14 Latin letters do have this property, and effectively the space is divided nearly evenly, and the resulting reward is close to the maximum. The same could be done with the next steps, provided a feature could be selected. If, however, the probability distribution is highly non-uniform, the aforementioned approach of matching some letters first could be more informative. Since the whole process is fully automatic, it can't be said a priori what concept structure would emerge and what features would be extracted and selected.

## 4  Binary Space Partitioning

The idea of binary space partitioning is depicted in Fig. 6. The input to a concept is a state space consisting of feature vectors of some multidimensional space. No matter how many inputs a concept has, we will consider all of them as a single vector. A hyperplane, depicted as a straight line, divides the input space into two disjoint subsets. We will call them positive and negative examples.

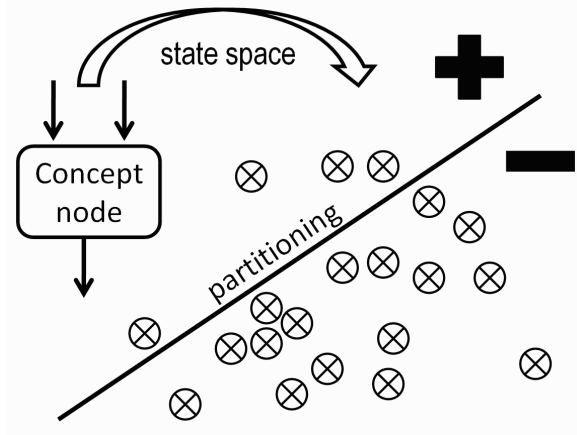

**Fig. 6.** Binary space partitioning.

Let us assume that $N_{pos}$ and $N_{neg}$ are the numbers of positive and negative examples observed, respectively, and $N_{pos} + N_{neg} > 0$. Then the probability of a positive outcome may be computed as:

$$P(\omega_{pos}) = \frac{N_{pos}}{N_{pos} + N_{neg}} \quad (42)$$

Fig. 7. presents the machine code implementation of the space partitioning. The instruction set of the virtual machine contains 65 unique codes, including conditional jumps, `RET` and `EXIT`. Both `RET` and `EXIT` terminate execution and quit, however, `RET` is interpreted as a positive outcome while `EXIT` as negative. The program gen-

erator of the heuristic search assures that RET and EXIT are placed in branches separated by a conditional jump.

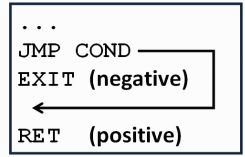

**Fig. 7.** Machine code implementation of space partitioning

Little can be said about the shape and the nature of the hyperplane that will be carried out by a codelet. Since the power of the program is virtually equivalent to that of a Universal Turing Machine, so is the structure of the hyperplane. The output may be either a feature vector, with the whole spectrum of values, or a binary flag. What matters is that, if the executed concept doesn't have internal states (e.g. no local static memory)[2], the partitioning really applies to the input state space.

Yet another problem may be encountered. With the growing $N_{pos} + N_{neg}$ the resulting probability would become insensitive to the incoming vectors. This problem has been solved by averaging the calculations over a window of 1000 recent vectors. As an interesting consequence observed on experiments, the candidate concepts exhibiting volatile probabilities do occasionally fall in low reward areas and consequently are dropped, while the stable concepts are preferred.

## 5  Reinforcement Learning

For every concept, the next actions to be taken at time $t$ is $a_{i,t+1}$ where $i=1,...,N$, N is the number of actions currently available. Since, however, actions may be removed and new actions added, the maximum number is virtually unlimited.

Both (13) and (14) TD-learning rules have been experimented with.

$$Q_{i,t} = Q_{i,t} + \alpha \left[ r_{i,t+1} + \gamma \max_{a_{i,t+1}} Q_{i,t+1} - Q_{i,t} \right] \qquad (13)$$

$$Q_{i,t} = Q_{i,t} + \alpha \left[ r_{i,t+1} + \gamma \overline{V}_{i,t+1} - Q_{i,t} \right] \qquad (54)$$

where $Q_{i,t}$ is the value of $a_i$ at time $t$, $\alpha$ is the learning rate, $\gamma$ is the discount factor, $r_{i,t+1}$ is the immediate reward at time $t+1$, and

$$\overline{V} = \frac{\sum_j p_{j,t} Q_{j,t}}{\sum_j p_{j,t}} \qquad (15)$$

where $p_{j,t}$ is the $P(\omega_{pos})$ at time $t$. It is must not be confused with the probability of selecting action $a_i$ at time $t$, which is calculated as

$$P(a_{i,t}) = \frac{Q_{i,t}}{\sum_j Q_{j,t}} \qquad (16)$$

---

[2] The virtual machine design allows for the concepts to have local static memory and internal states, and there are instruction codes that access the local memory. The discussion, however, is beyond the scope of this paper.

The future reward may be much larger than immediate reward and consequently may secure even a non-rewarding concept, like one performing a truncating function with no EXIT instruction.

The probability of adding a new action, possibly replacing the least rewarding one, here referred to as *exploration*, is

$$P_{\exp} \propto \frac{Q_{const}}{Q_{const} + \sum_i Q_i} \qquad (17)$$

where $Q_{const}$ is a predefined constant. The experiments with physical robot NAO have shown that most of the creative activity happens near the leaf nodes, while the area of stable concepts continuously extends from the root towards the higher levels.

## 6   Discussion

The application of immediate reward as the only source of reward, calculated locally at concept level and interpreted as intrinsic reward, does not exclude other more global fitness functions of intrinsic nature. In fact, it is already the parameters that control adding and removing actions, and the parameters that control the learning rate and discount factor, that may be understood as intrinsic motivation. Equation (17) may be also interpreted as the motivation towards curiosity, towards exploring the unknown areas.

Another question that has not been discussed above, and could also be contemplated in terms of intrinsic motivation, is the *artificial economics*. As was once mentioned, the computational overhead of evaluating the actions is also taken into account. A codelet that partitions the space optimally, if computationally too expensive, will be dropped as not rewarding  The individual actions compete according to (16). In the current implementation, however, the reward, as defined in (2), is divided by the (averaged) execution time. It is the execution on a virtual machine that enables precise resources management. Consequently, our goal of maximizing self-information gain in time may be redefined. Now the time may be understood as the execution time of the virtual machine code.

**References**

1. Deci, E.L., Ryan, R.M.: Intrinsic motivation and self-determination in human behavior. Plenum Press, New York (1985).
2. Oudeyer, P.-Y., Kaplan, F.: How can we define intrinsic motivation? Proceedings of the Eighth International Conference on Epigenetic Robotics: Modeling Cognitive Development in Robotic Systems, (2008).
3. Schmidhuber, J.: Formal theory of creativity, fun, and intrinsic motivation (1990-2010). IEEE Transactions of Autonomous Mental Development, vol. 2, no. 3. pp. 230-247. (2010).
4. Skaba, W.: Heuristic Search in Program Space for the AGINAO Cognitive Architecture. AGI-2011 Self-Programming Workshop, (2001). `http://www.iiim.is/wp/wp-content/uploads/2011/05/skaba-agisp-2011.pdf`
5. Skaba, W.: The AGINAO Self-Programming Engine. Submitted to: Journal of Artificial General Intelligence, Special Issue on Self-Programming, (2012). `http://aginao.com/pub/The_AGINAO_Self-Programming_Engine.pdf`
6. Goertzel, B.: The Hidden Pattern: A Patternist Philosophy of Mind. BrownWalker Press, (2006).
7. Hawkins, J., George, D.: Hierarchical temporal memory, (2006). `http://www.numenta.com/htm-overview/education/Numenta_HTM_Concepts.pdf`
8. Reiss, S.: Why Extrinsic Motivation Doesn't Exist. Psychology Today, (2011). `http://www.psychologytoday.com/blog/who-we-are/201108/why-extrinsic-motivation-doesnt-exist`
9. Cover, T.M., Thomas, J.A.: Elements of Information Theory. John Wiley & Sons, Inc. (1991). p. 20.
10. Tribus, M.: Thermodynamics and Thermostatics: An Introduction to Energy, Information and States of Matter, with Engineering Applications. Princeton, N.J., Van Nostrand, (1961).
11. Schmidhuber, J., Storck, J., Hochreiter, J.: Reinforcement Driven Information Acquisition In Non-Deterministic Environments. ICANN'95, vol. 2, pp. 159-164. EC2 & CIE, Paris. (1995).